\documentclass[letterpaper, 10 pt, conference]{ieeeconf}

\IEEEoverridecommandlockouts

\usepackage{graphics} 
\usepackage{epsfig} 
\usepackage{mathptmx} 
\usepackage{times} 
\usepackage{amsmath} 
\usepackage{amssymb}
\usepackage{kotex}
\usepackage{multirow}
\usepackage{booktabs}
\usepackage{color}
\usepackage{comment}
\usepackage[noadjust]{cite}
\usepackage{hyperref}
\hypersetup{ colorlinks=true, linkcolor=black, filecolor=magenta, urlcolor=magenta, }

\definecolor{green}{RGB}{31,173,0}

\title{\LARGE \bf
Part-Aware Data Augmentation for 3D Object Detection in Point Cloud*
}

\author{Jaeseok Choi$^{1}$, Yeji Song$^{1}$ and Nojun Kwak$^{1\dagger}$%
\thanks{*This work was supported by NRF grant (2021R1A2C3006659) and IITP grant (No. 2021-0-01343, Artificial Intelligence Graduate School Program), both funded by Korean Government. ($^{\dagger}$Corresponding Author: N. Kwak)}%
\thanks{$^{1}$J. Choi, Y. Song and N. Kwak are with Graduate School of Convergence Science and Technology, Seoul National University, Seoul 08826, South Korea. {\tt\footnotesize \{jaeseok.choi, ldynx, nojunk\}@snu.ac.kr}}%
}

\begin{document}

\maketitle
\thispagestyle{empty}
\pagestyle{empty}

%%%%%%%%%%%%%%%%%%%%%%%%%%%%%%%%%%%%%%%%%%%%%%%%%%%%%%%%%%%%%%%%%%%%%%%%%%%%%%%%
\begin{abstract}
Data augmentation has greatly contributed to improving the performance in image recognition tasks, and a lot of related studies have been conducted. However, data augmentation on 3D point cloud data has not been much explored. 3D label has more sophisticated and rich structural information than the 2D label, so it enables more diverse and effective data augmentation. In this paper, we propose part-aware data augmentation (PA-AUG) that can better utilize rich information of 3D label to enhance the performance of 3D object detectors. PA-AUG divides objects into partitions and stochastically applies five augmentation methods to each local region. It is compatible with existing point cloud data augmentation methods and can be used universally regardless of the detector's architecture. PA-AUG has improved the performance of state-of-the-art 3D object detector for all classes of the KITTI dataset and has the equivalent effect of increasing the train data by about 2.5$\times$. We also show that PA-AUG not only increases performance for a given dataset but also is robust to corrupted data. The code is available at \url{https://github.com/sky77764/pa-aug.pytorch}

\end{abstract}

%%%%%%%%%%%%%%%%%%%%%%%%%%%%%%%%%%%%%%%%%%%%%%%%%%%%%%%%%%%%%%%%%%%%%%%%%%%%%%%%
\section{INTRODUCTION}

3D object detection is critical for real-world applications such as in autonomous driving car and robotics.  Although 3D object detection research has been largely 
conducted, most of the works focus on architectures suitable for 3D point clouds \cite{lang2019pointpillars,yang2019std,he2020structure,Wang2019FrustumCS,Liu2020TANetR3,shi2020pv}.

Meanwhile, data augmentation plays an important role in boosting the performance of 3D models. 3D labeling is much more time-consuming compared to 2D labeling, which leads to most of the 3D datasets having a limited amount of training samples. Yet, 3D data augmentation has not been much explored.

\begin{figure}[t] 
    \begin{center}
        \includegraphics[width=1.0\linewidth]{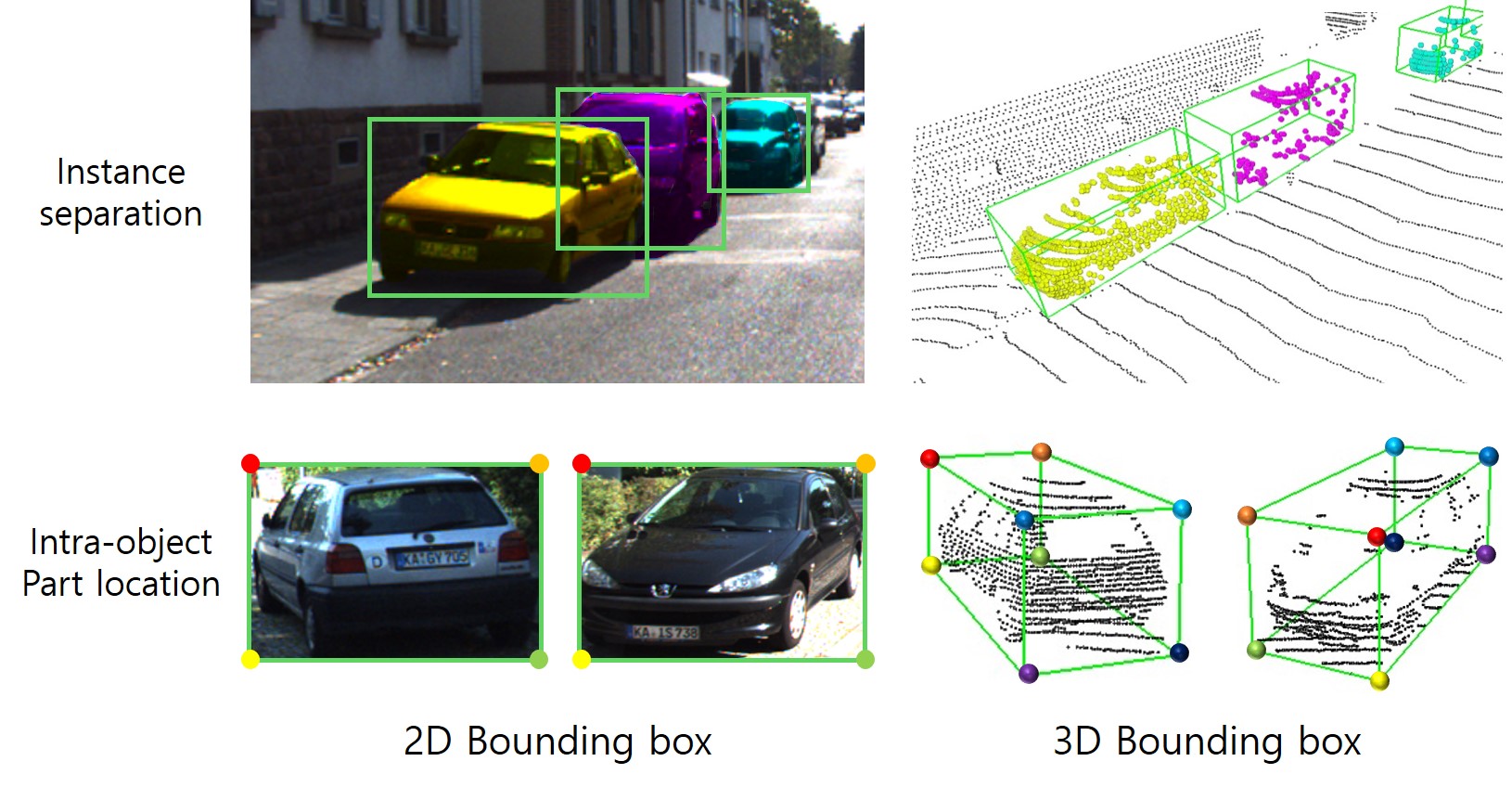}
    \end{center}
     \vspace{-2mm}
    \caption{\textbf{Comparison between 2D and 3D bounding box.} \textbf{Top - Instance separation:} Unlike 2D, 3D has separate instances in the box and rarely contains background points. \textbf{Bottom - Intra-object Part location:} Unlike 2D, the corners in 3D boxes can be assigned to a specific order using the heading direction of the object and using this order the information of the part location of the object can be obtained (different color represents different corners).}
    \label{fig:3d_bbox}
\end{figure}

Many works in 3D object detection apply data augmentation, such as translation, random flipping, shifting, scaling and rotation, directly extending typical 2D augmentation methods to 3D 
\cite{shi2020pv,Wang2019FrustumCS,chen2019object,hahner2020quantifying}. These existing methods are effective in improving performance. However, they did not fully utilize the 3D information. 3D ground-truth boxes have much richer structural information compared to 2D ground-truth boxes as they perfectly fit the object along with each direction. For example, the 2D label may contain other instances and background in the box, so the provided information contains much noise. On the other hand, 3D boxes provide sufficient information of a single object that is even occluded and have little background noise (Fig. \ref{fig:3d_bbox}, First row). Also, since the 2D boxes have no structural information about the objects, they cannot tell which part of the car is the `wheel'. However, we can be aware the wheels are located near the bottom corners using the intra-object part location information of 3D boxes (Fig. \ref{fig:3d_bbox}, Second row). Utilizing the unique characteristics of 3D boxes enables more sophisticated and effective augmentation which 2D augmentation cannot do.

In this paper, we propose a part-aware data augmentation method robust to various extreme environments by using structural information of 3D ground-truth boxes. The network can be aware of intra-object relation as it learns individual variation in an intra-object part. Our part-aware augmentation divides 3D ground-truth boxes into 8 or 4 partitions depending on the object type. It stochastically applies five augmentation methods to each partition, such as internal points dropout, cutmix \cite{yun2019cutmix}, cutmixup \cite{yoo2020rethinking}, sparse sampling, and random noise generation. The internal points dropout removes partitions stochastically and leaves the corner of an object. It enables the network to find the entire box when only some parts of the object are given. Cutmix and cutmixup respectively replace and mix points in the partition with other points from the same class and same partition location, which give the network a regularization effect. Sparse sampling samples point clouds from a dense partition, sparsifying the partition from which the network can learn more information of distant objects. Random noise generation allows the network to learn situations of severe occlusion.

Note that \cite{yun2019cutmix,yoo2020rethinking} apply cutmix and cutmixup to an image region with a patch from another class that the network could learn a relation across examples of different classes. In our work, however, points from the same class are mixed to give a regularization effect for intra-class examples. This reflects the task characteristics of 3D object detection that requires accurate localization while classifying 3 to 23 classes \cite{geiger2013vision, sun2019scalability, caesar2020nuscenes} centered on car, pedestrian and cyclist compared to \cite{yun2019cutmix,yoo2020rethinking} which classify 1000 classes of ImageNet.

Our proposed part-aware data augmentation improves KITTI \cite{geiger2013vision} Cyclist 3D AP of the PointPillars baseline \cite{lang2019pointpillars} up to 8.91\%p. Also, part-aware data augmentation enables the model to be robust in poor but inevitable environments, such as severe occlusion, low resolution, and inaccuracy due to snow or rain. In those situations, our work shows much less drop in accuracy than the baseline. In addition, part-aware augmentation performs well when data is insufficient, which has the equivalent effect of increasing the train data by about 2.5$\times$. As our work divides 3D box according to its structure and applies augmentation methods individually on the partitions, multiple augmentation methods are allowed to be used simultaneously without interference with each other. This can enhance the regularization effect a lot.

Our main contributions are:
\begin{itemize}
    \item As well as a partitioning method utilizing the structural information of 3D labels, we propose five partition-based 3D LiDAR data augmentation methods which significantly enhance performance when they are used together.

    \item Our work is compatible with existing LiDAR data augmentation methods and boosts conventional detectors’ performance with negligible cost.

    \item We show that proposed part-aware augmentation not only improves the recognition accuracy of given datasets but also obtains the robustness to corrupted data. 
\end{itemize}

%%%%%%%%%%%%%%%%%%%%%%%%%%%%%%%%%%%%%%%%%%%%%%%%%%%%%%%%%%%%%%%%%%%%%%%%%%%%%%%%

\begin{figure*}[t]
\vspace{2mm}
    \begin{center}
        \includegraphics[width=0.95\linewidth]{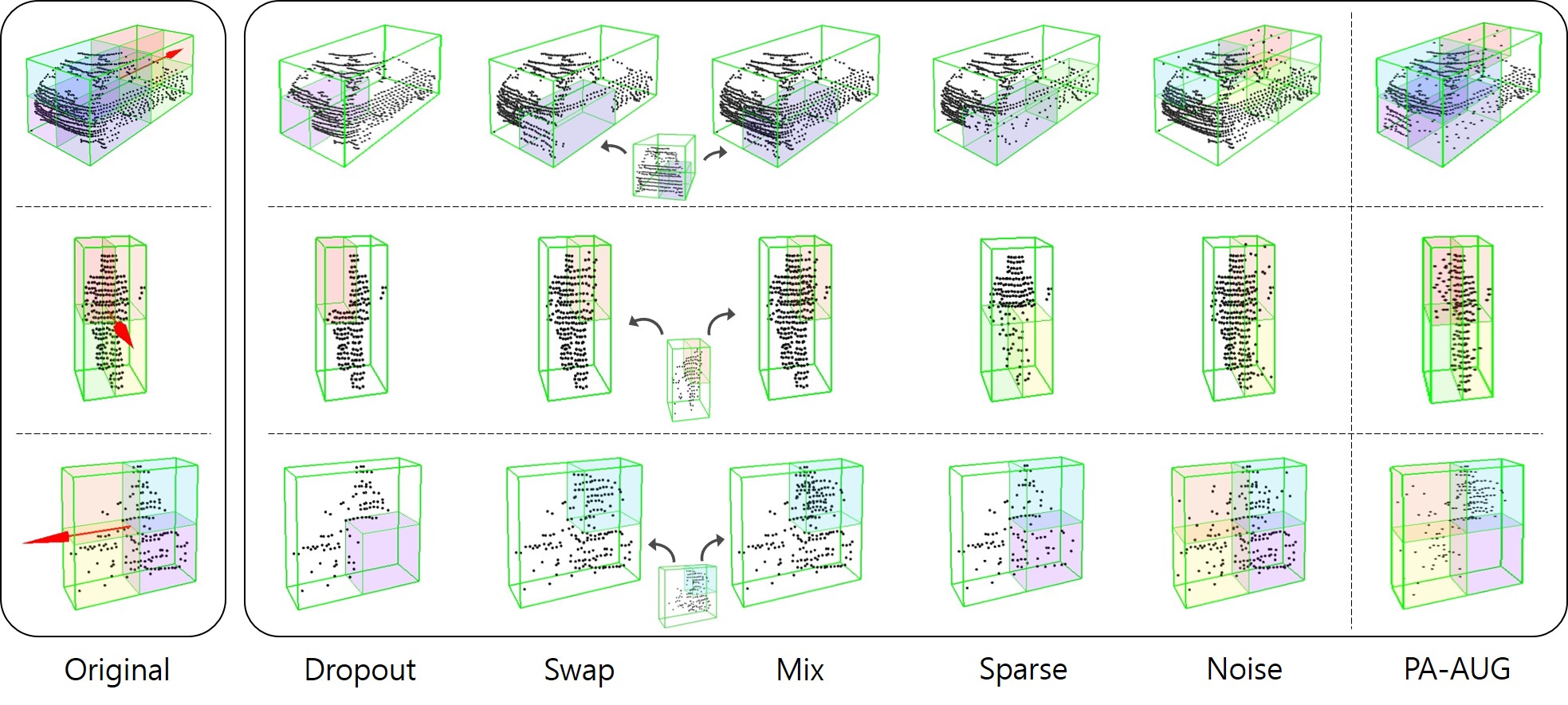}
    \end{center}
    \vspace{-4mm}
    \caption{\textbf{Part-aware partitioning and augmentation methods.} 
    The first column shows the original point cloud and part-aware partitioning method for Car, Pedestrian, and Cyclist classes. It divides the objects into 8, 4, and 4 partitions for each class. The other columns show examples of the proposed five partition-based augmentation methods and PA-AUG. The augmented partitions are marked with colors. Because Swap and Mix operations fetch points from different instances, the imported objects are also shown together.}
    \label{fig:methods}
   %\vspace{-1mm}
\end{figure*}

\section{RELATED WORKS}
\subsection{3D Object Detection}
Although RGB and LiDAR data can be used for 3D object detection, recent state-of-the-art (SOTA) detectors \cite{he2020structure, shi2020pv} rely only on LiDAR data. LiDAR-based 3D object detectors are largely classified into the projection, voxelization, and raw point cloud methods depending on the method for encoding point cloud.

The projection-based detection methods project point cloud data in the form of FV (Front View) or BEV (Bird Eye View) to use 2D convolutions. PIXOR~\cite{yang2018pixor} proposed a proposal-free, single-stage detector that uses BEV. LaserNet~\cite{meyer2019lasernet} proposed a method of predicting boxes in the form of distribution using FV. Since projection-based detectors use 2D CNNs, they have a great advantage in recognition speed, but their recognition performance is somewhat insufficient due to information loss that occurs in the projection process.

Voxelization-based methods quantize point cloud and encode them in a 3D matrix form to use 3D convolution. VoxelNet \cite{zhou2018voxelnet} divides the space into a 3D grid, combines the points included in each grid with fully-connected layers, and performs 3D convolution to regress 3D boxes. However, 3D convolution is very time-consuming. To resolve this problem, SECOND \cite{yan2018second} introduced sparse convolution, which greatly improved the detection speed of VoxelNet.

Unlike the projection and voxelization-based methods, the methods based on raw point cloud have no information loss of input. PointNet \cite{qi2017pointnet} and PointNet++ \cite{qi2017pointnet++} perform classification and segmentation by learning 3D representation of points using fully connected layers. PointRCNN \cite{shi2019pointrcnn} proposed a method which makes proposals using PointNet++ and refines 3D boxes with PointNet.

In recent years, many studies have been conducted to combine the advantages of the methods introduced above. PointPillars \cite{lang2019pointpillars} proposed a method of encoding a point cloud by voxelization in the form of a BEV 2D grid, significantly improving the detection speed. Part-$A^2$ \cite{shi2020points} Net creates proposals using raw point clouds to reduce the region of interest and then performs Box Refinement using voxelization. In addition, Part-$A^2$ Net proposed a method of using intra-object part location information of 3D labels. PV-RCNN \cite{shi2020pv} performs region proposal using voxelization and combines multi-scale voxel features with voxel set abstraction module to compensate for inaccurate recognition due to insufficient spatial resolution of voxelization-based proposals, greatly improving detection performance. SA-SSD \cite{he2020structure} also converts point cloud into a tensor using quantization and then extracts feature with 3D convolution. Also, to supplement the inaccurate detection due to downsampling, they proposed an auxiliary network that learns raw point cloud at a point level. Networks using these fusion methods currently show the best performance in LiDAR-based 3D Object Detection.

\subsection{2D Data Augmentation}
It has been demonstrated that data augmentation leads to gains in 2D image tasks such as classification and object detection \cite{zhong2020random, inoue2018data, taylor2018improving}. Especially, patch-based data augmentation methods that utilize patches cut and pasted among training images boosted performance. Image patches are zeroed-out in \cite{devries2017improved}, which encourages the model to utilize the full context of the image, on the other hand, deleted regions become uninformative. Cutmix \cite{yun2019cutmix} replaces deleted regions with a patch from another image and maximizes the proportion of informative pixels. These methods, when applied to CIFAR and ImageNet datasets, greatly improve the performance. Such improvements were also shown in low-level vision tasks. Cutblur \cite{yoo2020rethinking} cuts a low-resolution patch and replaces it with the corresponding high-resolution image region and vice versa. It has the same effect as making the image partially sparse and enables the model to learn both ``how" and ``where” when super-resolves the image.

In our work, the 2D image patch is extended to 3D partition. Using the 3D partition, we extend cutout \cite{devries2017improved}, cutmix \cite{yun2019cutmix}, and cutblur \cite{yoo2020rethinking} to 3D point clouds. Five proposed augmentation methods are simultaneously applied to the partitions which gives robustness to the model and significantly improves performance.

\subsection{3D Data Augmentation}
Considering the limited size of datasets for 3D object detection including KITTI datasets, data augmentation is one of the ways to alleviate overfitting and boost performance. The works \cite{shi2020pv, Wang2019FrustumCS, chen2019object} which showed the improved performance on 3D object detection adopted data augmentation methods such as translation, random flipping, shifting, scaling and rotation. Oversampling was also used to address foreground-background class imbalance problem \cite{yan2018second, shi2020pv, hahner2020quantifying}.

Despite their effectiveness on the models, existing data augmentation methods do not fully utilize richer information of point clouds compared to the counterparts for 2D images. We propose part-aware data augmentation which takes full advantage of spatial information unique in 3D datasets. 

Recently, automated data augmentation approaches have been actively studied. \cite{cheng2020improving} narrowed down the search space with an evolutionary-based search algorithm and adopted the best parameters discovered. \cite{li2020pointaugment} jointly optimized augmentor and classifier via an adversarial learning strategy. These approaches could be incorporated with our proposed part-aware data augmentation to further enhance the performance.

%%%%%%%%%%%%%%%%%%%%%%%%%%%%%%%%%%%%%%%%%%%%%%%%%%%%%%%%%%%%%%%%%%%%%%%%%%%%%%%%
\section{METHODS}
We propose a part-aware partitioning method that divides the object into partitions according to intra-object part location to fully utilize the structural information of 3D label. The term `intra-object part location' used in \cite{shi2020points} means a relative location of the 3D foreground points with respect to the corresponding bounding boxes and exploring the object part location improves performance. Part-aware partitioning is necessary to separate the characteristic sub-parts of an object and it enables more diverse and efficient augmentation than existing methods. Because the location of characteristic parts for each class is different, Car, Pedestrian and Cyclist are divided into 8, 4 and 4 partitions respectively (Fig. \ref{fig:methods}, First column). When using partition-based augmentation, instead of applying the same augmentation to the entire object, different augmentations can be applied to each intra-object sub-parts.

Point cloud $PC$ can be expressed by the union of foreground points $FP$ and background points $BP$ as below:
\begin{equation}
%\begin{aligned}
    PC = FP  \cup BP 
    %\end{aligned}
%\label{equ:pcl_def}
\end{equation}
\begin{equation}
 FP  = \cup^{N}_{i=1}B^{(i)} ,  
 \quad 
 B^{(i)} = \cup^{T}_{j=1}P^{(i)}_{j},
\end{equation}
where $B$ is the points in a 3D box, and $N$ is the number of boxes in a scene. $P$ is the internal points in a partition, and $T$ is the number of partitions in the box.

The set of augmented foreground points $FP_{aug}$ can be represented as 
\begin{equation}
%\begin{aligned}
   FP_{aug} = \cup^{N}_{i=1}\hat{B}^{(i)}, \quad
   \hat{B}^{(i)} = \cup^{T}_{j=1}\hat{P}^{(i)}_{j}. 
%\end{aligned}
\label{equ:fg_redef}
\end{equation}
Here, the bounding boxes and the partitions to which augmentation is applied are denoted as $\hat{B}$ and $\hat{P}$ respectively.

\subsection{Dropout Partition}
Dropout \cite{srivastava2014dropout} was first used in the feature-level to increase the regularization effect of the network by randomly making the activation of some nodes zero. After that, it was shown that dropout could be effectively applied to the input in the 2D image classification task \cite{devries2017improved}. Inspired by the previous works, we propose a partition-based dropout method that can be used effectively in 3D point clouds as below.

\begin{equation}
     \hat{B}^{(i)} = 
      \begin{cases}
          B^{(i)} & \textnormal{if}\ r_{i} = 0\\
          \cup^{T}_{j \neq d}P^{(i)}_{j} & \textnormal{if}\ r_{i} = 1
      \end{cases}  
      \quad \text{where}\ r_{i} \sim Ber(p_{D}). 
\label{equ:dropout}
\end{equation}

In Eq. (\ref{equ:dropout}), $Ber(\cdot)$ indicates Bernoulli distribution, and dropout is applied to each bounding box with a probability of $p_{D}$. The index $d$ indicates a randomly selected dropout partition among $T$ partitions. Dropout using a predefined partition can remove characteristic sub-parts of an object, making learning more robust.

\subsection{Swap Partition}
CutMix \cite{yun2019cutmix}, which is used in 2D image recognition, proposed an augmentation method that swaps random regions extracted from training samples. It can be applied to different classes by mixing class labels and has been shown effective for regularization. Inspired by the work, we propose a swap partition operation that utilizes intra-object part location information of 3D labels. The difference from CutMix is that our method swaps partitions of the same class and the same location in an object as follows.

\begin{equation}
\begin{split}
     \hat{B}^{(i)} =& 
      \begin{cases}
          B^{(i)}& \textnormal{if}\ r_{i} = 0\\
          \cup^{T}_{j\neq k}P^{(i)}_{j} \cup \hat{P}^{(i)}_{k} & \textnormal{if}\ r_{i} = 1 \textnormal{ and } \mid P^{(i)}_{k}\mid \neq 0
      \end{cases} \\ 
      & \qquad \qquad \qquad \ \text{where}\ r_{i} \sim Ber(p_{W}), 1 \leq k \leq T .
\label{equ:swap}
\end{split}
\end{equation}

\begin{equation}
      \hat{P}^{(i)}_{k} = P^{(i'->i)}_k 
\label{equ:swap2}
\end{equation}
\vspace{-4mm}
\begin{equation}
    P^{(i'->i)}_k = AffineTransform_{i}(P^{(i')}_k)
\label{equ:swap_trans}
\end{equation}

for $i \neq i'$, $1 \leq i' \leq N$ and $\mid P^{(i')}_k \mid \neq 0$.

As in the Eq. (\ref{equ:swap}) - (\ref{equ:swap_trans}), after selecting a box $i$ to swap with a probability of $p_{W}$ for all boxes, we swap a randomly selected non-empty $k^{th}$ partition in box $i$ with the $k^{th}$ partition in box $i'$. When swapping partitions, the partitions of different boxes have different scales, directions, and locations. So after transforming $P^{(i')}_k$ to the canonical coordinate system, we resize it to the scale of $P^{(i)}_k$ and transform it back to the coordinate system of $P^{(i)}_k$. As a result, $P^{(i'->i)}_k$ is created and the $k^{th}$ partition in box $i$ is replaced with it.

Because CutMix swaps patches of random areas, object can be swapped to the background area. And it could have a bad effect on learning. However, our partition-aware swap has no such problem and maximizes the effect of intra-class regularization by swapping only between the same class.

\begin{table*}[ht]
\vspace{2mm}
\begin{center}
\caption{\textbf{Performance comparison on the KITTI-val set.} The results are the average values of three repeated experiments. }
\begin{tabular}{l|ccc|ccc|ccc} \toprule
\multirow{2}{*}{Method}  & \multicolumn{3}{c|}{Car 3D (IoU=0.7)} & \multicolumn{3}{c|}{Cyclist 3D (IoU=0.5)} & \multicolumn{3}{c}{Pedestrian 3D (IoU=0.5)} \\
                         & Easy      & Mod.      & Hard      & Easy         & Mod.        & Hard        & Easy        & Mod.       & Hard       \\ \hline \hline
PointPillars \cite{lang2019pointpillars}             &     80.29      &    68.68       &    66.59       &        61.97      &       40.75      &        38.49     &      54.47       &       49.48     &     45.38       \\
PointPillars + Dropout   &     80.89      &    72.23       &    68.03       &       66.00       &       44.19      &    41.89         &     55.10        &      50.38      &    45.63        \\
PointPillars + Swap    &    81.45       &   68.60        &  66.85         &     66.66         &     44.94        &      42.62       &       56.92      &    51.97        &      47.32      \\
PointPillars + Mix     &    81.79       &   70.21        &  67.87         &     62.78         &     40.45        &      38.42       &       \textbf{59.98}      &    \textbf{54.60}        &      \textbf{48.87}      \\
PointPillars + Sparse      &    82.56       &    69.83       &    67.27       &     66.88         &     44.37        &      42.00       &   58.47          &    53.62        &  48.64          \\
PointPillars + Noise       &    82.03       &   68.37        &      65.81     &     66.44         &     44.31        &      41.79       &  57.81           &    52.55        &      47.73      \\
PointPillars + PA-AUG & \textbf{83.70}    &  	\textbf{72.48}    & 	\textbf{68.23}      &        \textbf{70.88}       &    \textbf{47.58}         &      \textbf{44.80}    &    57.38     &       51.85     &     46.91       \\ \hline
PV-RCNN \cite{shi2020pv}              &      89.15     &    80.43       &   78.48        &      85.54        &      71.21       &       65.42      &    66.08         &     59.48       &   55.22         \\
PV-RCNN + PA-AUG  &    \textbf{89.38}       &    \textbf{80.90}       &      \textbf{78.95}    &      \textbf{86.56}        &      \textbf{72.21}       &       \textbf{68.01}      &    \textbf{67.57}         &     \textbf{60.61}       &  \textbf{56.58}         \\

\bottomrule
\end{tabular}
\label{tab:KITTI_val}
\end{center}
\vspace{-4mm}
\end{table*}

\subsection{Mix Partition}
CutMixup \cite{yoo2020rethinking}, a combination of CutMix \cite{yun2019cutmix} and Mixup \cite{zhang2018mixup}, blends random areas of the training images, which is a quite effective data augmentation method in the task of image super-resolution. We applied it to our partition-based augmentation and call it Mix partition.
The detailed method is almost identical to the Swap partition except that Eq. (\ref{equ:swap2}) is replaced by
\begin{equation} \hat{P}^{(i)}_{k} = P^{(i)}_{k} \cup P^{(i'->i)}_k
\end{equation}
for $i \neq i'$, $1 \leq i' \leq N$, $\mid P^{(i')}_k \mid \neq 0$ and $r_{i} \sim Ber(p_{M})$.

The partition to mix is selected in the same way as the Swap operation. Likewise, the same partition transformation process is applied. The only difference is that it merges $P^{(i)}_{k}$ and $P^{(i'->i)}_k$ when creating augmented partition $\hat{P}^{(i)}_{k}$ rather than $P^{(i)}_{k}$ is replaced by $P^{(i'->i)}_k$.

\subsection{Sparsify Partition}
The density of LiDAR points decreases cubically as the distance of the box increases. As the point density decreases, the shape of the object cannot be fully recognized, which is one of the most significant factors in reducing the performance of LiDAR-based detectors. We propose sparsifying partitions as an augmentation method which makes some dense partitions sparse to improve distant objects' recognition.
The detail is as the following.

\begin{equation}
\begin{split}
     \hat{P}^{(i)}_{j} =& 
      \begin{cases}
          P^{(i)}_{j}& \textnormal{if}\ r_{j} = 0\\
          S^{(i)}_{j}& \textnormal{if}\ r_{j} = 1\textnormal{ and } \mid P^{(i)}_{j}\mid > C_{S}
      \end{cases} \\ 
      & \qquad \qquad \qquad \ \text{where}\ r_{j} \sim Ber(p_{S}).
\label{equ:sparse}
\end{split}
\end{equation}

As in Eq. (\ref{equ:sparse}), it selects partitions to augment with the probability of $p_{S}$ among the dense partitions with the number of points over $C_{S}$. Then, $C_{S}$ points of the partition are sampled using Farthest Point Sampling (FPS) and it is denoted as $S^{(i)}_{j} \subset P_j^{(i)}$.

\subsection{Add Noise to Partition}
Since the RGB-image-based detectors are greatly influenced by the illuminance of the surrounding environment, the augmentation methods that change the brightness and color help improve performance. Likewise, LiDAR-based detectors are vulnerable to weather changes such as rain or snow that can cause noise and occlusion in point cloud data. We propose a partition-based augmentation method to be robust to noise caused by various reasons as follows:
\begin{equation}
%\begin{aligned}
     \hat{P}^{(i)}_{j} = 
      \begin{cases}
          P^{(i)}_{j}& \textnormal{if}\ r_{j} = 0\\
          P^{(i)}_{j} \cup P_{noise}& \textnormal{if}\ r_{j} = 1
      \end{cases} \quad
      \text{where}\ r_{j} \sim Ber(p_{N})
\label{equ:noise}
%\end{aligned}
\end{equation}

As in Eq. (\ref{equ:noise}), it selects partitions to augment with the probability of $p_{N}$. Then, it adds randomly generated $C_{N}$ noise points $P_{noise}$ to the selected partition $P_j^{(i)}$. 

\subsection{PA-AUG}
The five augmentation methods using part-aware partitioning introduced above can be used individually, but because the methods are independent, different augmentation methods can be applied to an object multiple times. Therefore various combination of augmentations can be created, applying each operation independently so that different operations can be applied to one partition. However, if all augmentations are used without a specific order, interference may occur between operations. In order to minimize this, we apply Dropout-Swap-Mix-Sparse-Noise in order. We call it \textbf{PA-AUG}, which stochastically applies five operations. It can take advantage of each and show a strong regularization effect.

%%%%%%%%%%%%%%%%%%%%%%%%%%%%%%%%%%%%%%%%%%%%%%%%%%%%%%%%%%%%%%%%%%%%%%%%%%%%%%%%
\section{EXPERIMENTS}

\begin{table}[t]
\setlength{\tabcolsep}{1.5mm}
\begin{center}
\caption{\textbf{Parameters used in KITTI experiments.}}
% \footnotesize
\scriptsize
\begin{tabular}{l|ccccccc} \toprule
\multirow{2}{*}{Method} & \multicolumn{7}{c}{Parameters} \\
                          & $p_{D}$ & $p_{W}$ & $p_{M}$ & $C_{S}$ & $p_{S}$ & $C_{N}$ & $p_{N}$ \\ \hline
                        
Dropout                   & 1.0/0.3  & -  & -  & -  & -  & -  & -  \\
Swap                      &  - &  1.0/0.7 & -  & -  & -  &  - &  - \\
Mix                       & -  &  - &  0.3/1.0 & -  & -  &  - &  - \\
Sparse                   &  - &  - & -  & 40/50  & 0.3/0.3  & -  & -  \\
Noise                     &  - &  - & -  & -  &  - & 5/10  & 0.3/0.1  \\
PA-AUG                    & 0.2/0.2  & 0.2/0.2  & 0.2/0.2  & 40/40  & 0.1/0.1  & 10/10  & 0.1/0.1 \\ \bottomrule
\end{tabular} 
\label{tab:KITTI_parameters}
\end{center}
\vspace{-6mm}
\end{table}

\subsection{KITTI}  \label{section:KITTI}
\textbf{Settings}
The KITTI object detection benchmark dataset \cite{geiger2013vision} consists of 7,481 training samples and 7,518 testing samples. In order to verify the effectiveness of PA-AUG, we separated the training dataset into 3,712 training samples and 3,769 validation samples \cite{chen20153d}. Since our augmentation methods are applied stochastically, we report the average values of 3 repeated experiments in Table \ref{tab:KITTI_val}.

PointPillars \cite{lang2019pointpillars} uses two separate networks for Car and Cyclist/Pedestrian classes. We use a batch size of 2 for Car network and 1 for Cyc/Ped network. And we train 160 epochs for Car and 80 epochs for Cyc/Ped network.\footnote{default parameters in https://github.com/traveller59/second.pytorch} PV-RCNN \cite{shi2020pv} uses a single network for all classes. We train the network with batch size 8 for 80 epochs. The parameters of the proposed augmentation methods are shown in Table \ref{tab:KITTI_parameters}. The left values of `/' are parameters of the Car network, and the right values are parameters of the Cyc/Ped network. Basic data augmentations such as ground-truth oversampling \cite{yan2018second}, rotation, translation, and flipping are used before applying our partition-based augmentations. For other parameters not mentioned, the settings of each original paper are used. 

\textbf{Results}
Table \ref{tab:KITTI_val} shows the effect of each partition-based augmentation methods and PA-AUG. Precision and recall curves are computed using 11 points. All the proposed standalone augmentation methods performed better than the baseline algorithms without our data augmentation (PointPillars \cite{lang2019pointpillars} and PV-RCNN \cite{shi2020pv}) in most cases. We have found that the cases in which each operation significantly increases are different. For example, Dropout does not improve the Easy score of Car a lot, but it does for Mod. and Hard cases. Other operations, on the contrary, increase the Easy score a lot compared to Mod. and Hard scores. For the Cyc/Ped network, Mix operation achieves the highest scores for Pedestrian class, but scores for Cyclist class are remarkably low. Interestingly, PA-AUG achieves the highest performance improvement on average through even improvements for all scores, which means the proposed partition-based augmentations have synergy effects when used together. Also, PA-AUG  improves all the scores of PV-RCNN \cite{shi2020pv}, one of the current state-of-the-art LiDAR-based detectors.

\begin{table}[t]
\setlength{\tabcolsep}{3pt}
\begin{center}
\caption{\textbf{Robustness Test.} The 3D AP$_{Hard}$(IoU=0.7) on KITTI-val are reported. The values in parentheses are the performance decrease of each corrupted dataset compared to KITTI-val. The baseline model is PointPillars \cite{lang2019pointpillars}.}
\begin{tabular}{l|cccc}   \toprule
\multirow{2}{*}{Augmentation} & \multicolumn{4}{c}{Dataset}                                                             \\
                              & \multicolumn{1}{c}{KITTI} & \multicolumn{1}{c}{KITTI-D} &      \multicolumn{1}{c}{KITTI-S} & \multicolumn{1}{c}{KITTI-J} \\ \hline
Baseline & 67.35 & 58.91(-8.44) & 56.89(-10.46) & 56.66(-10.69) \\
+ Dropout & \textbf{68.05} & \textbf{64.09}(\textcolor{green}{\textbf{-3.96}}) &  62.27(\textcolor{green}{-5.78}) &
57.11(\textcolor{red}{-10.94}) \\
+ Swap & 66.69 & 60.84(\textcolor{green}{-5.85}) & 59.55(\textcolor{green}{-7.14}) &
54.90(\textcolor{red}{-11.79})\\
+ Mix & 67.91 & 63.13(\textcolor{green}{-4.78}) & 63.42(\textcolor{green}{-4.49}) & 56.03(\textcolor{red}{-11.88}) \\
+ Sparse &  67.59   & 62.73(\textcolor{green}{-4.86}) & 62.90(\textcolor{green}{-4.69}) & 40.02(\textcolor{red}{-27.57}) \\
+ Noise & 65.99      & 58.67(\textcolor{green}{-7.32}) & 59.64(\textcolor{green}{-6.35}) & \textbf{58.25}(\textcolor{green}{\textbf{-7.74}}) \\
+ PA-AUG & 67.74      & 63.61(\textcolor{green}{-4.13}) & \textbf{64.20}(\textcolor{green}{\textbf{-3.54}}) & 57.91(\textcolor{green}{-9.83}) \\
\bottomrule
\end{tabular}   
\label{tab:robustness}
\end{center}
\vspace{-6mm}
\end{table}

\subsection{Robustness Test}    \label{section:robustness}
\begin{figure}[t] 
    \vspace{2mm}
    \begin{center}
        \includegraphics[width=1.0\linewidth]{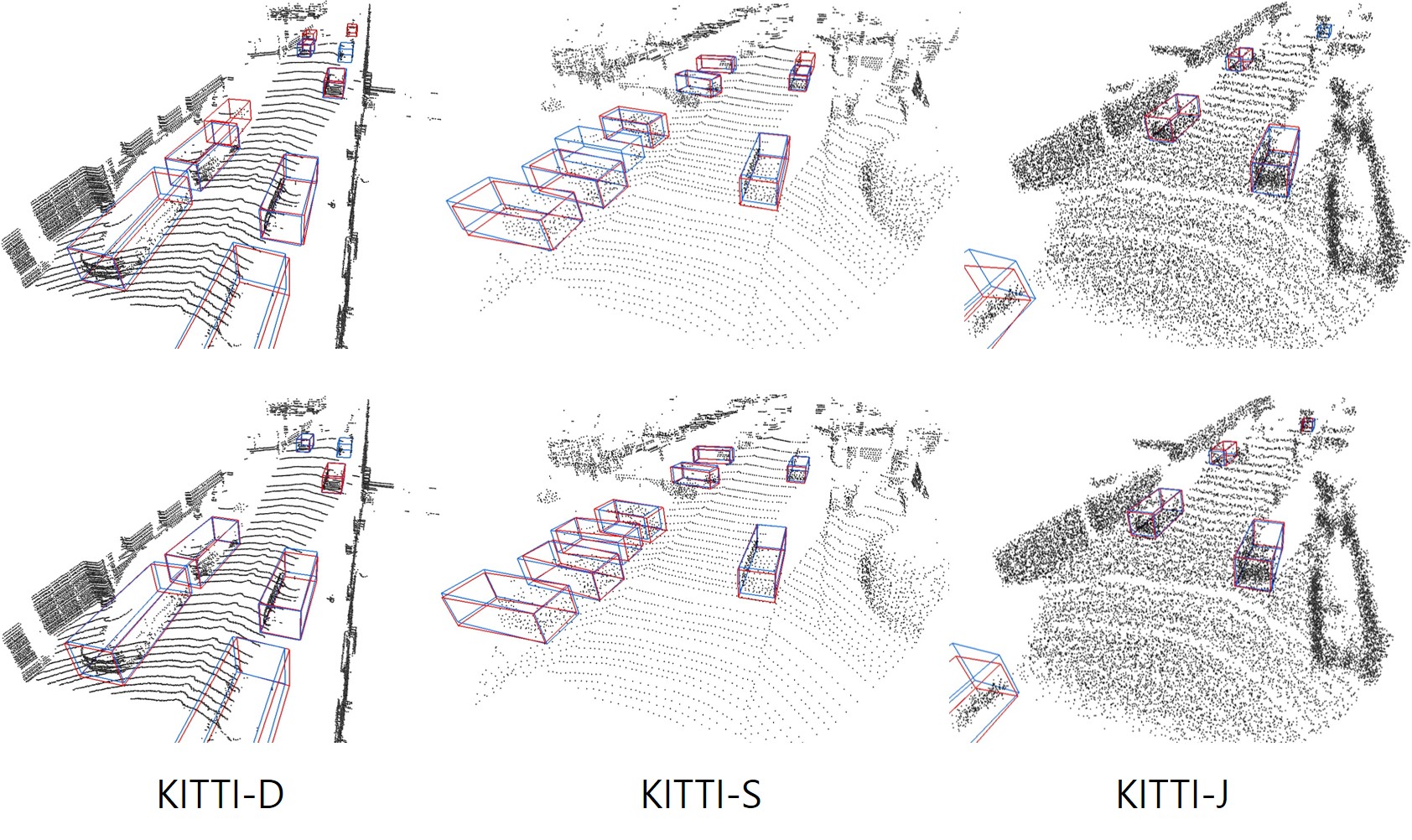}
    \end{center}
    \vspace{-2mm}
    \caption{\textbf{Qualitative results on corrupted KITTI datasets.} The upper row shows the PointPillars results, and the lower row shows the PointPillars + PA-AUG results. The ground-truth and predicted boxes are shown in blue and red, respectively. 
    }
    \label{fig:qualitative}
    %\vspace{-2mm}
\end{figure}

\textbf{Settings} 
We evaluate the robustness of our proposed augmentations using three corrupted KITTI-\textit{val} datasets, KITTI-D, KITTI-S, and KITTI-J. KITTI-D (Dropout) is a dataset in which some of the foreground points are removed by dropping out a portion of all objects. For fairness, not making it the same as the dropout used for our proposed augmentation, a random dense area with many points is selected for the part to be dropped out. KITTI-S (Sparse) is a dataset that leaves only 30\% of points using Farthest Point Sampling (FPS) across the point cloud. Finally, KITTI-J (Jittering) is a dataset that adds Gaussian noise $X \sim \mathcal{N}(0,0.1^2)$ for all points. Each corrupted dataset is designed to closely simulate the actual scenario of the cases when occlusion is severe, LiDAR has a low resolution, or LiDAR is incorrect. Some examples with detection results are shown in Fig. \ref{fig:qualitative}.

\textbf{Results} 
In Table \ref{tab:robustness}, the 3D AP$_{Hard}$(IoU=0.7) scores on the KITTI-\textit{val} and its corrupted datasets are reported. The values in parentheses are the performance decrease of each corrupted dataset compared to KITTI-\textit{val} (leftmost). In the table, the best performance on each dataset is denoted in bold. Dropout, Swap, Mix, and Sparse operations all showed less performance decrease on the KITTI-D and KITTI-S datasets than the baseline. However, the performance decreased significantly on the KITTI-J dataset. On the other hand, Noise operation showed a smaller decrease than the baseline on every corrupted dataset. PA-AUG takes advantage of each operation evenly and shows the most robust performance for corrupted datasets.

\begin{table}[ht]
%\normalsize
\begin{center}
\caption{\textbf{Comparison of partitioning methods.}}
\begin{tabular}{l|ccc}  \toprule
\multirow{2}{*}{Method} & \multicolumn{3}{c}{KITTI-\textit{val} 3D AP$_{Easy}$}                                                                            \\
                           & Car@0.7                 & \multicolumn{1}{c}{Cyclist@0.5} & \multicolumn{1}{c}{Pedestrian@0.5} \\ \hline
Random                     & \multicolumn{1}{c}{80.64} & 62.42                             & 55.60                                \\
Part-aware                 & \multicolumn{1}{c}{\textbf{83.70}} & \textbf{70.88}                             & \textbf{57.38}                          \\
\bottomrule
\end{tabular}
\label{tab:partitioning}
\end{center}
\vspace{-3mm}
\end{table}

% \subsection{Partitioning Method}              \label{section:partitioning}
\subsection{Ablation Study}              \label{section:ablation}
\textbf{Partitioning Method} 
To verify the need for the part-aware partitioning method, we randomly create partitions without part information. The random partitions are created with the same number and the same direction as the part-aware partitions, but the scales and positions are randomly generated for each object. We apply the proposed partition-based augmentations equally to the random partitions and the part-aware partitions. As shown in Table \ref{tab:partitioning}, random partition-based augmentation has significantly less performance improvement than part-aware partition for all classes. From this result, it can be seen that the part information plays a critical role in applying the proposed partition-based augmentations.

\begin{table}[ht]
\begin{center}
\caption{\textbf{Comparison of the number of partitions.}}
\begin{tabular}{c|ccc} \toprule
\multirow{2}{*}{\# Partitions} & \multicolumn{3}{c}{KITTI-\textit{val} 3D AP$_{Easy}$}                                                       \\
                               & Car@0.7                         & Cyclist@0.5        & Pedestrian@0.5     \\
                               \hline
2                              &    80.85                               & 68.81                     &    57.14                  \\
4                              &    80.67                               & \textbf{70.88}       & \textbf{57.38}       \\
8                              & \multicolumn{1}{c}{\textbf{83.70}} & \multicolumn{1}{c}{67.87} & \multicolumn{1}{c}{52.94} \\
\bottomrule
\end{tabular}
\label{tab:num_partitions}
\end{center}
\end{table}

\textbf{The Number of Partitions}
Since we roughly know the location of the characteristic sub-parts for each class, we defined a different number of partitions for each class using this prior knowledge. In order to check whether the number of partitions defined is actually the most effective, we experiment varying the number of partitions. As shown in Table \ref{tab:num_partitions}, the best performance can be achieved by using 8, 4 and 4 partitions for Car, Cyclist and Pedestrian classes. From these results, it can be seen that if the number of partitions is too large or too small, the effect on the original data becomes too small or too large, thus weakening the regularization effect and reducing the performance.

\subsection{Data Efficiency}    \label{section:efficiency}

\begin{figure}[t] 
\vspace{2mm}
    \begin{center}
        \includegraphics[width=1.0\linewidth]{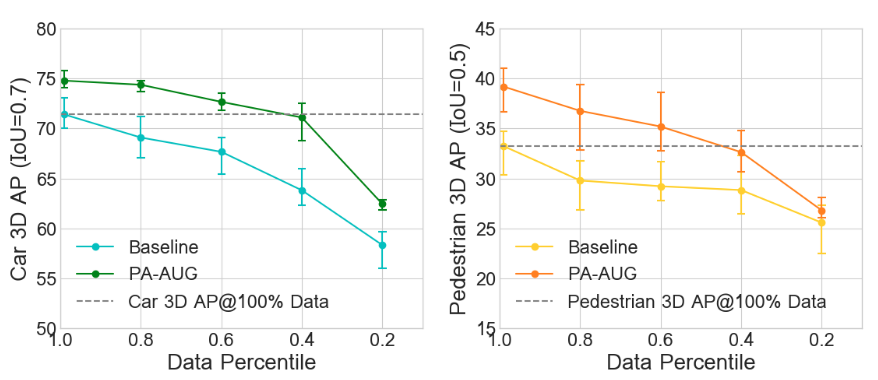}
    \end{center}
    \vspace{-2mm}
    \caption{\textbf{Data Efficiency Test.}
    The graphs show the 3D AP$_{Easy}$ scores on KITTI-\textit{val} according to the size of the training data subsets. All other data augmentations except PA-AUG are not used.}
    \label{fig:efficiency}
    \vspace{-2mm}
\end{figure}

We downsample the KITTI datasets, taking subsets with number of 20\%, 40\%, 60\%, 80\% training examples to verify how PA-AUG performs under very little data. Green and orange dots  in Fig. \ref{fig:efficiency} show the performance of PA-AUG with the full datasets and four subsets, respectively indicating Car and Pedestrian categories. Cyan and yellow dots in Fig. \ref{fig:efficiency} show the results of the baselines. In these experiments, all other data augmentations except PA-AUG are not used to verify the effectiveness of PA-AUG alone. 
The results show that PA-AUG is effective not only in the full dataset, but also in data subsets. PA-AUG using only 40\% of examples achieves 3D AP comparable with the baselines using full datasets in both Car and Pedestrian. That is, PA-AUG is about 2.5$\times$ more data-efficient.

We notice that the performance drop in PA-AUG is steeper than the baseline as the size of the datasets decreases. This phenomenon is due to the relative lack of information of original objects in PA-AUG since modified and augmented datasets are provided where the original data itself is highly insufficient. The performance drop may slow down when smaller augmentation parameters are applied. Even so, PA-AUG shows the higher performances in full datasets and all subsets than the baseline since the improvement is much more significant.

%%%%%%%%%%%%%%%%%%%%%%%%%%%%%%%%%%%%%%%%%%%%%%%%%%%%%%%%%%%%%%%%%%%%%%%%%%%%%%%%
\section{CONCLUSIONS}
We have presented PA-AUG which makes better use of 3D information of point clouds than the conventional methods. We divide the objects into 8 or 4 partitions according to intra-object part location and apply five separate augmentation methods which can be used simultaneously in a partition-based way. The proposed data augmentation methods can be universally applied to any architecture, and PA-AUG further improves one of the SOTA detectors on KITTI dataset. Experimental results show that PA-AUG can improve robustness to corrupted data and enhance data efficiency. Because of the generality of the proposed methods, we believe that it can be used in any tasks utilizing 3D point clouds such as semantic segmentation and object tracking.
However, there are some limitations in applying PA-AUG to other 3D datasets. The bounding box must be oriented and objects must not overlap too much. These limitations make it difficult to apply our method to indoor datasets such as ScanNet \cite{dai2017scannet}.


%%%%%%%%%%%%%%%%%%%%%%%%%%%%%%%%%%%%%%%%%%%%%%%%%%%%%%%%%%%%%%%%%%%%%%%%%%%%%%%%

%%%%%%%%%%%%%%%%%%%%%%%%%%%%%%%%%%%%%%%%%%%%%%%%%%%%%%%%%%%%%%%%%%%%%%%%%%%%%%%%

\bibliographystyle{unsrt}
\bibliography{IEEEexample}

\end{document}